\documentclass[10pt,twocolumn,letterpaper]{article}

\usepackage[pagenumbers]{wacv} 

\usepackage{graphicx}
\usepackage{amsmath}
\usepackage{amssymb}
\usepackage{booktabs}
\usepackage{adjustbox, multicol, multirow}
\usepackage[title]{appendix}
\usepackage{ltablex, booktabs}
\usepackage{float, dblfloatfix}
\def\etal{\textit{et al.}}

\usepackage[pagebackref,breaklinks,colorlinks]{hyperref}

\usepackage[capitalize]{cleveref}
\crefname{section}{Sec.}{Secs.}
\Crefname{section}{Section}{Sections}
\Crefname{table}{Table}{Tables}
\crefname{table}{Tab.}{Tabs.}

\def\wacvPaperID{967} 
\def\confName{WACV}
\def\confYear{2025}

\begin{document}

\title{Keypoint Aware Masked Image Modelling}

\author{Madhava Krishna\\
IIIT Delhi\\
{\tt\small madhava20217@iiitd.ac.in}
\and
A V Subramanyam\\
IIIT Delhi\\
{\tt\small subramanyam@iiitd.ac.in}
}
\maketitle
\begin{abstract}
SimMIM is a widely used method for pretraining vision transformers using masked image modeling. However, despite its success in fine-tuning performance, it has been shown to perform sub-optimally when used for linear probing. We propose an efficient patch-wise weighting derived from keypoint features which captures the local information and provides better context during SimMIM's reconstruction phase. Our method, KAMIM, improves the top-1 linear probing accuracy from 16.12\% to 33.97\%, and finetuning accuracy from 76.78\% to 77.3\% when tested on the ImageNet-1K dataset with a ViT-B when trained for the same number of epochs. We conduct extensive testing on different datasets, keypoint extractors, and model architectures and observe that patch-wise weighting augments linear probing performance for larger pretraining datasets. We also analyze the learned representations of a ViT-B trained using KAMIM and observe that they behave similar to contrastive learning with regard to its behavior, with longer attention distances and homogenous self-attention across layers. Our code is publicly available at \href{https://github.com/madhava20217/KAMIM}{https://github.com/madhava20217/KAMIM}.
\end{abstract}

\section{Introduction}
\label{sec:intro}

Since their introduction, transformers \cite{vaswani2017attention} have become deeply ingrained in the practice of deep learning. Not only are they highly parallelizable and capable of handling longer sequences than recurrent networks, they also scale much better with increased data. With the introduction of the Vision Transformer (ViT) \cite{dosovitskiy2020image}, deep learning in vision has undergone a paradigm shift. In contrast with Convolutional Neural Networks (CNNs) \cite{lecun1989handwritten}, which inherently learn local features in the beginning and obtain a global view later, ViTs have been shown to learn global representations early on \cite{raghu2021vision}. However, they need much more data than CNNs to learn local representations, requiring supervised training on massive datasets like JFT300 to offer competitive performance.

\begin{figure}[hbt!]
\centering
{\includegraphics[width = \linewidth]{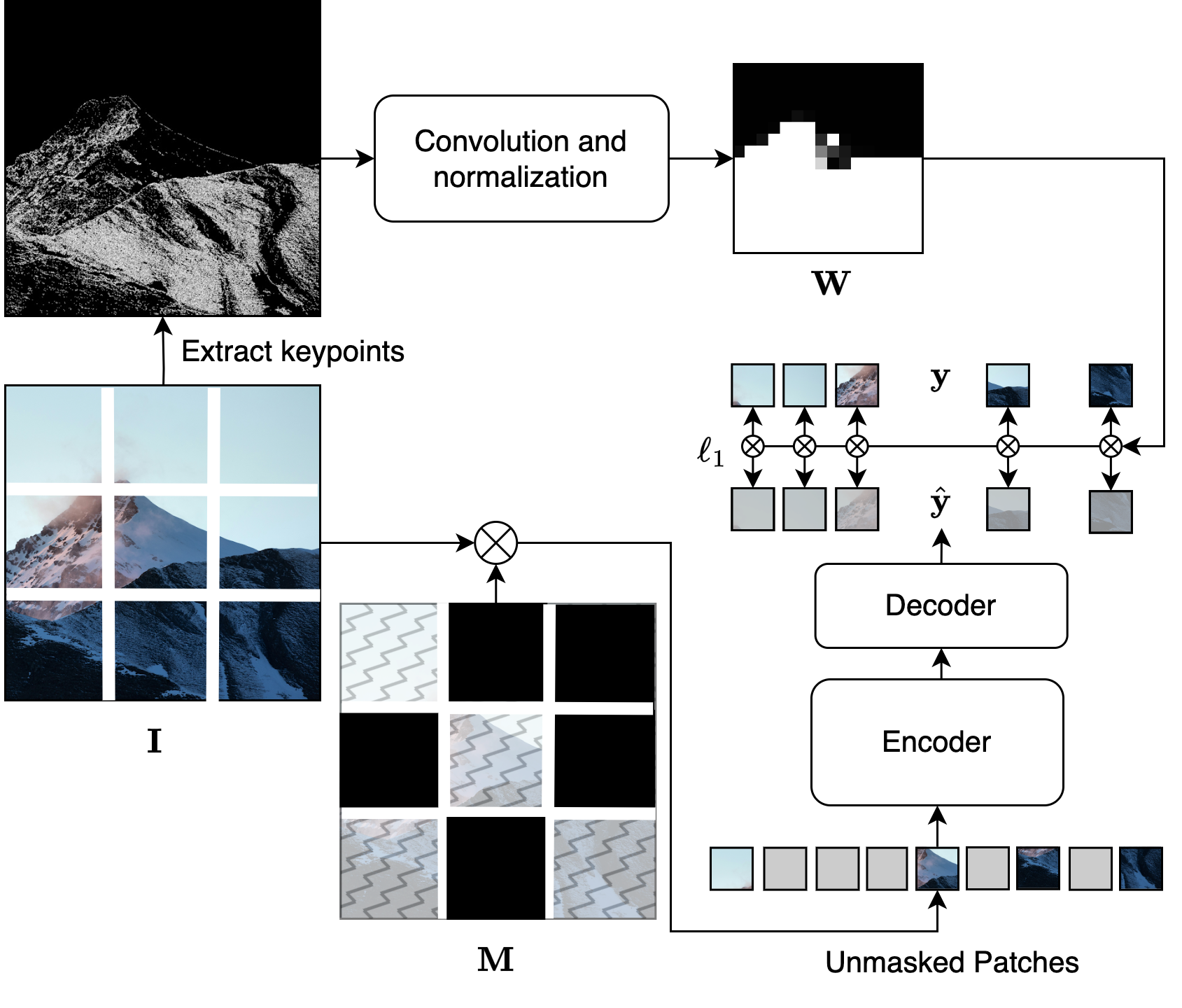}}
\caption{A diagram depicting the working of KAMIM. We calculate FAST keypoints of an image $\mathbf{I}$ and then compute the density of keypoints over patches of a pre-defined size ($w_{ps}$) to use as a weight $\mathbf{W}$ during the reconstruction phase. An averaging convolutional kernel is used to efficiently calculate the keypoint density and further obtain the weight matrix $\mathbf{W}$. In order to control the weighting factor, a temperature parameter is used.
Distinct from SimMIM, an $\ell_1$-loss with weight $\mathbf{W}$ is used on predicted pixel value from a lightweight prediction head.}
\label{fig:simmim}
\end{figure}

Pre-training has been of vital significance in training transformers, both in language and vision. Masked language modeling (MLM), used for training BERT models (Bidirectional Encoder Representations from Transformers) \cite{devlin2018bert}, serve as an efficient and performant self-supervised pre-training method. Masked image modeling (MIM) is the equivalent of MLM for vision and involves masking a part of an input image and learning representations by reconstructing it. SimMIM \cite{xie2022simmim} is one of many existing pre-training methods that utilizes MIM for efficiently learning good representations for downstream tasks. By resorting to a pixel-prediction pre-text task and using a lightweight prediction head, SimMIM has been used to successfully train billion-parameter Swin-v2 transformers \cite{liu2021swin, liu2022swin}. However, it fails to provide competitive results when tested for linear probing against other methods. 

We observe that SimMIM attends equally to all patches in the  reconstruction loss. We hypothesize that paying differential attention can benefit the training process. 
Towards this, 
we propose KAMIM wherein we introduce a weighted reconstruction loss. This weight is obtained from the density of keypoints in the input image. Thus, we can force the network to pay more attention if the region has lot of information and in turn dense keypoints. By default, the keypoints are detected by the FAST (Features from Accelerated Segment Test) algorithm \cite{rosten2006machine}, but we also experiment with SIFT \cite{lowe1999object} and ORB \cite{rublee2011orb}. We observe a twofold improvement in linear-probing accuracy when tested on ImageNet-1k \cite{deng2009imagenet} with a ViT-B, along with a 0.96\% increase in finetuning accuracy. We conduct tests with different hyperparameters affecting the keypoint density calculation and the extent of the weighting and report results for ViTs and Swin transformers of different parameter counts. We also test the performance of our method on CIFAR10, CIFAR100 \cite{krizhevsky2009learning}, iNaturalist \cite{van2018inaturalist}, and Places365 \cite{zhou2014learning} datasets and observe that KAMIM outperforms SimMIM in linear probing  while still providing strong results in finetuning if the pretraining is conducted on a large dataset like ImageNet. We then contrast KAMIM against SimMIM, establishing similarities of KAMIM to contrastive learning algorithms like MoCo \cite{he2020momentum}, which display a larger attention distance and more homogenous attention maps compared to masked image modeling.
\section{Related Works}
\label{sec:related_works}

\subsection{Self-supervised Learning}

Balestriero \etal \cite{balestriero2023cookbook} suggest that there exist four classes of self-supervised learning methods - the deep metric learning family, the self-distillation family, the canonical correlation analysis family, and the masked modeling family. The deep metric learning family consists of methods that maximize the similarity between different versions of the same input and covers techniques like contrastive learning \cite{chen2020simple, dwibedi2021little, yang2022mutual, yuan2021multimodal}. The self-distillation family \cite{grill2020bootstrap, naeem2023silc, lee2022self, oquab2023dinov2, fang2021seed, zhou2021ibot} aims to perform distillation on the learner by means of a target network. The canonical correlation analysis family \cite{bardes2021vicreg,zbontar2021barlow,caron2020unsupervised} maximizes the covariance between representations of differently augmented images. Masked modeling aims to reconstruct clean signals given a corrupted signal.

\subsection{Masked Image Modeling (MIM)} Masked image modeling works by masking parts of the input image and reconstructing them. Vincent \etal \cite{vincent2008extracting} denoised corrupted images to train autoencoders in a layer-wise fashion. Pathak \etal \cite{pathak2016context} used context encoders to learn representations in a self-supervised manner by inpainting. Since the introduction of transformers, a number of MIM methods have been developed -- masked modeling on image pixels \cite{bao2021beit, he2022masked, xie2022simmim} and other prediction targets like HOG \cite{dalal2005histograms} features \cite{wei2022masked}. Some methods, like iBOT \cite{zhou2021ibot} and DINOv2 \cite{oquab2023dinov2}, use MIM in conjunction with self-distillation.

Our work closely resembles Zhuge \etal \cite{zhuge2024patch} in that they identify the importance of the relative difference between patches during reconstruction. While they work on a similar problem, our approach fundamentally differs from \cite{zhuge2024patch}. Their primary contribution, PASS, consists of a module that provides a data sampling approach for selecting patches and weighs each patch differently based on how informative the patch is. They show that PASS can be used with SimMIM \cite{xie2022simmim} and MAE \cite{he2022masked} to outperform the base approaches while requiring less data.


\subsection{Keypoint Detection Algorithms}
Keypoint detection algorithms are used to find points of interest in a given image. Handcrafted keypoint detection algorithms like the Harris corner detector \cite{harris1988combined}, SIFT \cite{lowe1999object}, FAST \cite{rosten2006machine}, SURF \cite{bay2006surf}, and ORB \cite{rublee2011orb} use gradients, difference of Gaussians or difference in pixel intensities of an image to identify corners or blobs. More recent works, like DetNet \cite{lenc2016learning}, TILDE \cite{verdie2015tilde}, and LIFT \cite{yi2016lift} use convolutional neural networks to identify keypoints. Newer techniques such as SuperPoint \cite{detone2018superpoint} and TUSK \cite{jin2022tusk} learn to identify keypoints in a self-supervised manner, not requiring explicitly labelled data for this purpose.
\section{Proposed Method}
\label{sec:methodology}

We first give a brief description of SimMIM followed by the FAST algorithm for keypoint detection. We then introduce our proposed method.

\subsection{SimMIM}
Given an image $\mathbf{I}$ and a mask $\mathbf{M}$ comprising of boolean values for each patch of the image, we generate a masked view $\mathbf{I'}$ where patches at masked locations are replaced by a learnable value. Suppose ${P'}$ is the set of indices of patches that have been masked, and $\hat{\mathbf{y}}_{i, j}$ is the predicted value for pixels of patch at location $(i, j)$ having original value $\mathbf{y}_{i, j}$. Then, the objective we minimize for SimMIM is:

\begin{align}
    \mathcal{L} &= \frac{1}{\lvert P'\rvert} \sum\limits_{(i,j) \in P'} \lVert\hat{\mathbf{y}}_{i, j} - \mathbf{y}_{i, j}\rVert_1
\end{align}

where $|P'|$ is the cardinality of the set $P'$, and $\lVert.\rVert_1$ indicates the $\ell_1$ loss.

\subsection{FAST Algorithm}

The Features from Accelerated Segment Test (FAST) algorithm \cite{rosten2006machine} is a fast and computationally efficient algorithm for identifying corners from an image. For a point $p$ tested for being a corner, the FAST algorithm first computes the differences in intensity for a circle of 16 pixels around it. It first tests for 4 points in four cardinal directions around the point and then checks for other points if the difference is greater than a set threshold. The point is taken as a candidate corner if at least 12 contiguous points from the circle satisfy the constraints.

After identifying candidates, an ID3 \cite{quinlan1986induction} decision tree is used on the 16 pixels surrounding them to reduce false positives. Non-maximal suppression is then used to reduce the number of overlapping corners.


\subsection{Keypoint Aware SimMIM (KAMIM)}
We observe that SimMIM weighs each patch equally, paying no attention to the information contained within. Our hypothesis is that weighing each patch differently could improve the training process.
Additionally, we want to maintain the training efficiency of SimMIM while making minimal changes to the training process. Thus, we resort to a keypoint-based weighting mechanism, which, as per our experiments, adds little to no overhead to training time, while providing better global understanding.

\subsubsection{Setup}
Given an image $\mathbf{I}$ of size $d \times d$, we calculate FAST keypoints for $\mathbf{I}$ and plot them on a grid of the same dimension to obtain image $\mathbf{F}$. Each element in $\mathbf{F}$ has a value of either 0 or 1. Locations identified as keypoints will have a value of 1 while others will have a value of 0.

We then have two hyperparameters for KAMIM:
\begin{enumerate}
    \item Patch Size ($w_{ps}$): This determines over what patch size the keypoint density for an image is calculated. The density of keypoints is calculated for each $(w_{ps} \times w_{ps})$ patch in image $\mathbf{F}$. We then perform exponentiation and scaling to derive a weight for each patch such that the minimum possible weight is 1, corresponding to vanilla SimMIM's weighting.
    \item Temperature ($T$): The temperature inversely determines the extent of the impact of the weight. A higher temperature would place less importance on the weights derived from KAMIM and act closer to SimMIM, while a lower temperature would put more importance on them.
\end{enumerate}

We describe our training objective next.

\subsubsection{Implementation}
\label{sec:implementation}
 Given a patch size $w_{ps}$, we identify the density of key points in each $w_{ps}\times w_{ps}$ patch. This can be done efficiently by convolving a kernel $\mathbf{K}$ equal to an all-ones matrix scaled by $\frac{1}{w_{ps}^2}$ and stride in both axes equal to $w_{ps}$ over $\mathbf{F}$. From this step, we obtain a $\frac{d}{w_{ps}} \times \frac{d}{w_{ps}}$ keypoint density matrix, $\boldsymbol{\Omega}$, with each cell equal to the average number of key points in the corresponding patch of $\mathbf{I}$. 

To derive the final weights $\mathbf{W}$, we perform the following:
\begin{align}
    \label{eq:weight}
    \mathbf{W}_{i, j} &= \frac{e^{\mathbf{\Omega}_{i, j}/T}}{\min\limits_{i', j'} (e^{\mathbf{\Omega}_{i', j'}/T})}
\end{align}

where $\mathbf{W}_{i,j}$ is the weight of patch in position $(i, j)$. The denominator in equation \ref{eq:weight} ensures that the weight of any patch is greater than or equal to 1.

Finally, for boolean mask $\mathbf{M}$, set of masked indices of patches $P'$, original value of pixels in patch $(i,j)$ being $\mathbf{y}_{i,j}$ and predicted value $\hat{\mathbf{y}}_{i,j}$, we have:

\begin{align}
    \label{eq:wsimmim}
    \mathcal{L} &= \frac{1}{\lvert P'\rvert} \sum\limits_{(i,j) \in P'} \lVert \mathbf{W}_{i, j}(\hat{\mathbf{y}}_{i,j} - \mathbf{y}_{i,j})\rVert_1
\end{align}

In equation \ref{eq:wsimmim}, a higher weight is assigned to a patch which has a high density of keypoints. This allows the network to focus more on such patches and extract relevant information. In our experiments, we show that the networks trained with KAMIM approach shows superior performance.
\section{Results}
\label{sec:results}

\subsection{Experimental Setup}
For pre-training, we use an AdamW \cite{loshchilov2017decoupled} optimizer with weight decay of 0.05, $\beta_1 = 0.9$, $\beta_2 = 0.999$, and a learning rate of 8e-4. We employ a cosine learning rate scheduler for 100 epochs with 10 epochs of warmup. 

For fine-tuning and linear probing, we use the AdamW optimizer with the same values of weight decay, $\beta_1$, and $\beta_2$, but with a learning rate of 5e-3. A cosine LR scheduler was used for 100 epochs with 10 epochs of warmup. A single linear layer is used for prediction after average pooling and flattening embeddings from the backbone model. For linear probing with ViTs, we use the 8$^{th}$ layer embeddings as per \cite{bao2021beit}, while the last layer embeddings are used for Swin transformers. While the reported metrics for ViTs do not use layer normalization \cite{ba2016layer} before the linear head is applied, our experiments indicate that the obtained results hold true despite that.

\begin{table*}[hbt!]
\centering
\begin{tabular}{llllllllll}
\hline
\textbf{Model} &
\textbf{SimMIM (LP)} &
\textbf{KAMIM (LP)} &
\textbf{SimMIM (FT)} &
\textbf{KAMIM (FT)} &
\textbf{\#Params} \\ \hline
ViT-T  & 12.37& \textbf{13.75} & \textbf{70.49} & 70.41 & 5.5M  \\\hline
ViT-S  & 20.99 & \textbf{22.68} & 76.8 & \textbf{77.02} & 21.6M \\\hline
ViT-B  & 16.12 & \textbf{33.97} & 76.78 & \textbf{77.30} & 85.7M \\\hline
Swin-T & 14.37 & \textbf{14.53} & 77.94 & \textbf{78.12} & 27.5M \\\hline
Swin-B & \textbf{20.42} & 18.16 & 79.58 & \textbf{80.02} & 86.7M \\\hline
\end{tabular}
\caption{Top-1 accuracy of KAMIM and SimMIM on the ImageNet-1K dataset. LP indicates linear probing while FT denotes finetuning performance. The values of hyperparameters are $T 
= 0.25$ and $w_{ps} = 8$. The scores reported are for the validation split.}
\vspace{-0.4cm}
\label{tab:simmim_vs_wsimmim}
\end{table*}

\subsubsection{Data Augmentations}
For both SimMIM and KAMIM, we use the same boolean mask generator -- a masking patch size of $32 \times 32$ and a masking ratio of 0.6. For pre-training, a random resized crop is performed with dimension $192 \times 192$ with minimum and maximum ratios of 0.67 and 1, respectively, followed by random vertical and horizontal flips with $p = 0.5$.

For linear probing and fine-tuning, we use a resize operation first to convert the image to size $192 \times 192$, followed by TrivialAugment \cite{muller2021trivialaugment} with strength sampled from \{0, \dots, 30\}. After this, a random erasing operation is used with $p = 0.5$, the scale between 0.02 and 0.33, and the aspect ratio of the erased region between 0.3 and 3.3. Finally, color normalization is performed, followed by CutMix \cite{yun2019cutmix} or MixUp \cite{zhang2017mixup}, which is determined by an equiprobable random choice. 

\subsubsection{Models Used for the Experiments}
We experiment with ViTs and Swin transformers, conducting tests with ViT-T, ViT-S, ViT-B, Swin-T, and Swin-B to determine scalability. The parameter configuration for ViT-T and ViT-S are obtained from Touvron \etal \cite{touvron2021training} except with input resolution $192 \times 192$.

Our experiments suggest that the optimal temperature ($T$) and weight patch size ($w_{ps}$) is $0.25$ and $8$, respectively, with linear probing top-1 accuracy as the metric of choice.

\subsubsection{KAMIM Hyperparameters}
For all experiments comparing KAMIM to SimMIM in  \Cref{subsec:hyperparameter_sweep,subsec:simmim_vs_wsimmim,subsec:corner_detection_algos}, we use the best hyperparameters, that is, weight patch size $w_{ps} = 8$ and temperature $T = 0.25$.

\begin{table}[hbt!]
\centering
\begin{tabular}{llll}
\hline
${w_{ps}}$ & \textit{T} & \textbf{LP} & \textbf{FT} \\ \hline
\multirow{4}{*}{8}  & 0.1  & 32.97          & \textbf{77.59} \\ 
                    & 0.25 & \textbf{33.97} & 77.29          \\ 
                    & 0.5  & 31.95          & 77.11          \\ 
                    & 1    & 23.93          & 76.94          \\ \hline
\multirow{4}{*}{16} & 0.1  & 31.38          & 77.43          \\ 
                    & 0.25 & 33.05          & 77.48          \\ 
                    & 0.5  & 30.82          & 77.06          \\ 
                    & 1    & 29.51          & 77.13          \\ \hline
\multirow{4}{*}{32} & 0.1  & 25.71          & 77.09          \\ 
                    & 0.25 & 31.73          & 77.41          \\ 
                    & 0.5  & 30.91          & 77.41          \\ 
                    & 1    & 25.73          & 77.18          \\ \hline
\multicolumn{2}{c}{SimMIM} & 16.12 & 76.78 \\ \hline
\end{tabular}
\caption{KAMIM performance with different values of hyperparameters -- weight patch size and temperature. `FT' depicts fine-tuning performance while `LP' is for linear probing. Evaluation has been done on the pre-defined validation set of ImageNet-1k. We see that KAMIM consistently outperforms SimMIM over the entire range of hyperparameters tested for.}
\label{tab:wsimmim_parameter_variation}
\vspace{-0.4cm}
\end{table}

\subsection{Comparison with SimMIM}
\label{subsec:simmim_vs_wsimmim}
We conduct testing on ImageNet-1K \cite{deng2009imagenet} and obtain results as in \Cref{tab:simmim_vs_wsimmim}. We note that KAMIM consistently outperforms SimMIM in linear probing. The difference is significant for ViT-B, providing a 2.1x performance uplift in top-1 accuracy, increasing the top-1 accuracy from 16.12\% to 33.97\%. We observe that the differences are less pronounced for smaller models like ViT-T and ViT-S and more significant for larger ones. In addition to this, we observe competitive performance in fine-tuning, surpassing SimMIM for all the cases except ViT-T. We also see that in case of Swin-B, SimMIM shows better performance. This can be attributed to the fact that Swin transformers use hierarchical attention and local windows, which can exploit local context to build a better global understanding.

\subsection{Effect of Varying Hyperparameters}
\label{subsec:hyperparameter_sweep}
We conduct testing on the ImageNet-1k dataset with a ViT-B model. Results are given in 
\Cref{tab:wsimmim_parameter_variation}.
We observe the best results for linear probing with a weight patch size of 8 and a temperature of 0.25. A general trend we note is that lower values of temperatures, which increase the patch-wise weighting, improve the linear probing performance. $T = 1$ and $w_{ps} = 8$, results in a linear probing accuracy of 23.93\% which is $10.04\%$ lower than that obtained with $T = 0.25$. A similar trend can be seen for lowering the weight patch size, indicating that a more granular aggregation of KAMIM weights can result in better performance. This is seen in with $T = 0.25$ and $w_{ps} = 32$ which gives $2.24\%$ lower accuracy than $w_{ps} = 8$.

We note that while a higher value of $T$ doesn't perform as well compared to a lower value of $T$, it is still able to outperform SimMIM in linear probing by a significant margin.

\begin{table*}[t]
\centering
\resizebox{\textwidth}{!}{%
\begin{tabular}{lllllllllllllll}
\toprule
\multirow{2}{*}{\textbf{Algorithm}} &
  \multicolumn{2}{l}{\textbf{ImageNet}} &
  \multicolumn{2}{l}{\textbf{iNaturalist}} &
  \multicolumn{2}{l}{\textbf{Places365}} &
  \multicolumn{2}{l}{\textbf{CIFAR10}} &
  \multicolumn{2}{l}{\textbf{CIFAR100}} &
  \multicolumn{2}{l}{\textbf{\begin{tabular}[c] {@{}l@{}} CIFAR10 \\ (IN-1k PT)\end{tabular}}} &
  \multicolumn{2}{l}{\textbf{\begin{tabular}[c]{@{}l@{}}CIFAR100\\ (IN-1k PT)\end{tabular}}} \\
 &
  \textbf{LP} &
  \textbf{FT} &
  \textbf{LP} &
  \textbf{FT} &
  \textbf{LP} &
  \textbf{FT} &
  \textbf{LP} &
  \textbf{FT} &
  \textbf{LP} &
  \textbf{FT} &
  \textbf{LP} &
  \textbf{FT} &
  \textbf{LP} &
  \textbf{FT} \\ \midrule
SimMIM &
  16.12 &
  76.78 &
  6.92 &
  79.8 &
  26.04 &
  \textbf{53.2} &
  \textbf{62.83} &
  \textbf{94.13} &
  \textbf{37.46} &
  \textbf{78.92} &
  61.65 &
  97.03 &
  36.49 &
  84.58 \\
KAMIM &
  \textbf{33.97} &
  \textbf{77.30} &
  \textbf{17.60} &
  \textbf{80.48} &
  \textbf{40.47} &
  53.03 &
  59.01 &
  94.12 &
  35.67 &
  77.26 &
  \textbf{72.63} &
  \textbf{97.43} &
  \textbf{51.77} &
  \textbf{85.07} \\ \midrule
PASS (SimMIM)$^*$ & 
    --  &
  \textbf{82.1} &
  -- &
  -- &
  -- &
  -- &
  -- &
  -- &
  -- &
  -- &
  -- &
  97.1 &
  -- & 
  81.7 \\ \bottomrule
\end{tabular}
}
\caption{KAMIM compared against SimMIM on other datasets. Pre-training is conducted for 100 epochs with 10 epochs of warmup on the same dataset. For the CIFAR10 and CIFAR100 results with ImageNet-1k weights, we directly take ImageNet weights after 100 epochs of pre-training with 10 epochs of warmup. Finetuning on all datasets consists of 100 epochs with 10 epochs of warmup. The results for PASS are reported directly from the paper. Note that PASS$^*$ is pre-trained for 200 epochs on $224 \times 224$ images and a ViT patch size of 14.}
\label{tab:other_datasets}
\end{table*}

\subsection{Comparison with PASS}
PASS \cite{zhuge2024patch} pre-trains for 200 epochs on a $224 \times 224$ image resolution with a $14 \times 14$ ViT patch size, albeit on a smaller sampled dataset. Results are given in \Cref{tab:other_datasets}.

KAMIM does better than PASS-SimMIM when tested for finetuning on the CIFAR10 and CIFAR100 datasets. On CIFAR100, our method achieves 3.37\% higher accuracy despite training for considerably lesser number of epochs. However, KAMIM is unable to outperform PASS-SimMIM on ImageNet-1K, which we attribute to the larger number of epochs, smaller ViT patch size, and higher image resolution that PASS-SimMIM has been trained on.

\subsection{Results on Other Datasets}

We conduct testing with a ViT-B on the CIFAR-10, CIFAR-100 \cite{krizhevsky2009learning}, Places365 \cite{zhou2014learning} and iNaturalist-2021 \cite{van2018inaturalist} datasets. Pre-training is done on the same dataset as the evaluation. We choose the best hyperparameters as observed in \Cref{subsec:hyperparameter_sweep}. Results are given in \Cref{tab:other_datasets}.

For larger datasets like iNaturalist or the Places365, which contain 2.7M and 1.8M training images, respectively, we see KAMIM consistently beating SimMIM by significant margins. In linear probing for iNaturalist, KAMIM achieves 17.60\% accuracy, whereas, SimMIM obtains 6.92\%. We observe similar gain for Place365.

In case of smaller datasets like the CIFAR10 and CIFAR100, both containing 50,000 images, the results edge out in SimMIM's favor. When using a model pre-trained on a larger pretraining dataset like ImageNet, we observe that KAMIM significantly outperforms SimMIM on these smaller datasets. This indicates that the number of examples required for learning good representations is higher for KAMIM compared to the original method.

We also note that SimMIM gives similar results with or without ImageNet checkpoint weights. We can infer that the benefit of pretraining from a larger dataset such as ImageNet does not immediately transfer when using SimMIM. On the other hand, with ImageNet checkpoints, our method gives a boost of 13.62\% and 16.10\% on CIFAR-10 and CIFAR-100 respectively under linear probing. This indicates that our method converges much faster than SimMIM. 

\begin{table}[ht]
\centering
\resizebox{\linewidth}{!}{ 
\begin{tabular}{llll}
\hline
\textbf{Detector} & \textbf{LP} & \textbf{FT} & \textbf{Epoch Time} \\ \hline
None (SimMIM)     & 16.12                   & 76.78               & =1x                 \\ \hline
FAST              & \textbf{33.97}          & 77.29               & $\sim$1x            \\ \hline
SIFT              & 31.31                   & \textbf{77.37}      & $\sim$1.06x         \\ \hline
ORB               & 32.36                   & 77.17               & $\sim$1.07x        \\ \hline
\end{tabular}
}
\caption{Performance comparison for different keypoint detectors. Testing is done on ImageNet-1K with a ViT-B for 100 epochs with 10 epochs of warmup for each one out of pretraining, fine-tuning, and linear probing.}\label{tab:other_corner_detectors}
\vspace{-0.4cm}
\end{table}

\subsection{Corner Detection Algorithms}
\label{subsec:corner_detection_algos}
We replace the FAST keypoint detection algorithm with two other keypoint detectors -- SIFT \cite{lowe1999object}, and ORB \cite{rublee2011orb}. With the best hyperparameters seen in \Cref{subsec:hyperparameter_sweep}, we test on the ImageNet-1k dataset with a ViT-B. Results are given in \Cref{tab:other_corner_detectors}.

We observe that FAST features provide superior performance compared to other methods while also taking less time to train than other detectors. In addition, SIFT performs the best in fine-tuning but fails to outperform ORB or FAST in linear probing. All of the weighted methods perform better than SimMIM in linear probing.

\subsection{Performance in Reconstruction}
We gauge the performance of the two methods in reconstruction using their lightweight prediction heads with Peak Signal-to-Noise-Ratio (PSNR) and Structural Similarity Index Measure (SSIM) scores. We conduct testing on ImageNet's validation set with ViT-B models. We observe similar scores from both methods. SimMIM resulted in a PSNR of $12.058$ and SSIM of $0.1095$, while KAMIM gave a PSNR of $12.063$ and SSIM of $0.1076$. The tests are, so far, inconclusive to show the superiority of one method over the other for reconstruction. Some examples of input images and reconstructions are given in \cref{fig:reconstruction}.

\setcounter{figure}{2}
\begin{figure*}[b]
\includegraphics[width = \textwidth]{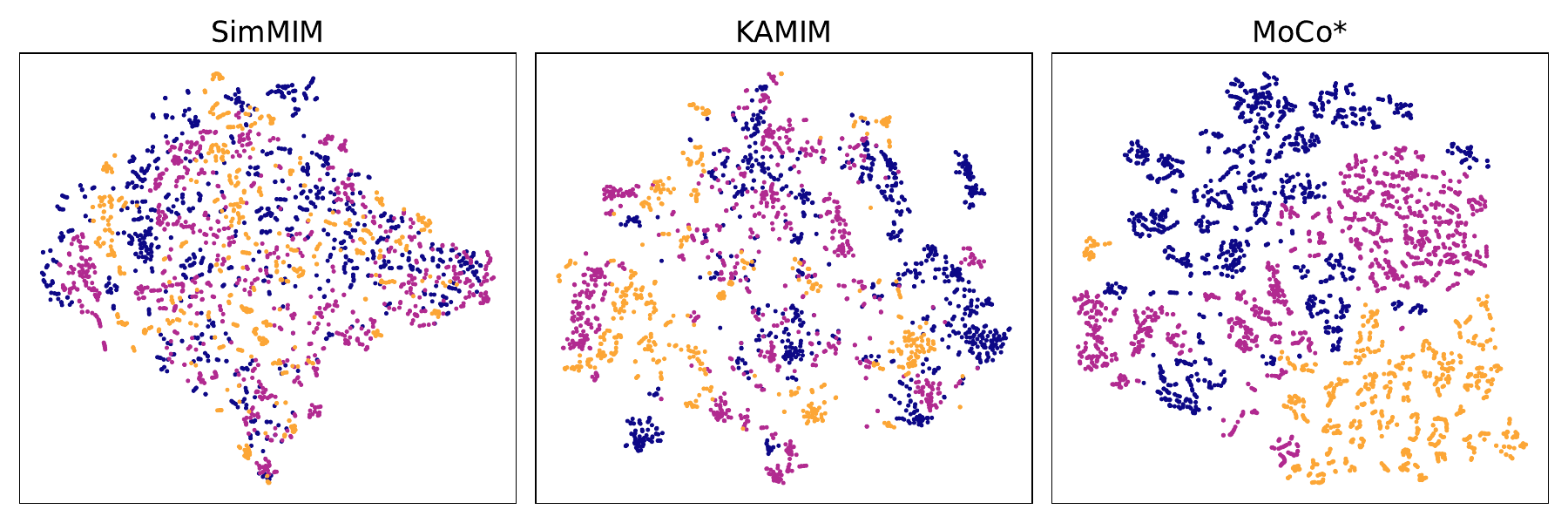}
\caption{The t-SNE visualization of a set of token-level representations of images from different classes represented in different colours. The embeddings from the last layer are used. Note that the \textit{cls} token is dropped, and only the 144 remaining tokens are used for SimMIM and KAMIM, and 196 tokens for MoCo.}
\label{fig:tsne_token_level}
\vspace{-0.3cm}
\end{figure*}

\setcounter{figure}{1}
\begin{figure}[t]
\includegraphics[width = \linewidth]{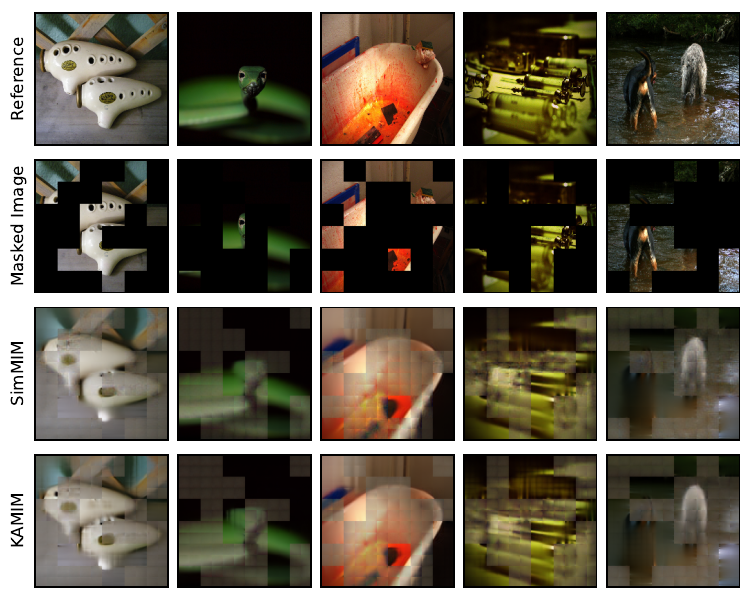}
\caption{Reconstructed images from SimMIM and KAMIM. The first row depicts the original image, and the second row contains the masked images. We unnormalize the reconstructed images to obtain these visualizations. We see that both methods are close in terms of visual fidelity and it is hard to judge which one is better. SimMIM sometimes performs better, as seen in image 5, while KAMIM does better at capturing details, as in image 1.}
\label{fig:reconstruction}
\end{figure}
\section{Behavior of Learnt Representations and Attention}

We refer to Park \etal \cite{park2023self} to compare the self-attentions and the representations learnt by a ViT-B trained on an image resolution of $192 \times 192$ with the two methods. To facilitate a better comparison with MoCo \cite{he2020momentum}, we also include results from their paper. Please note that all figures involving MoCo use their provided model checkpoint trained for 800 epochs on $224 \times 224$ images.

\setcounter{figure}{3}
\begin{figure}[b]
\centering
\includegraphics[width = 0.65\linewidth]{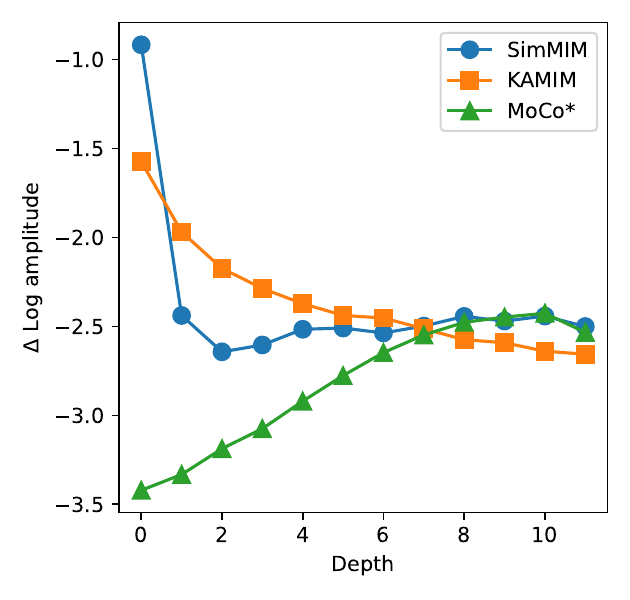}
\vspace{-0.5cm}
\caption{The plot resulting from the Fourier analysis of the output of intermediate layers in. The hidden states first undergo a 2D Fourier transform, followed by a log amplitude operation and differencing with the first latent to obtain a relative log amplitude plot.}
\label{fig:fourier_analysis}
\end{figure}

\begin{figure*}[t]
\centering
\includegraphics[width = 0.94\textwidth]{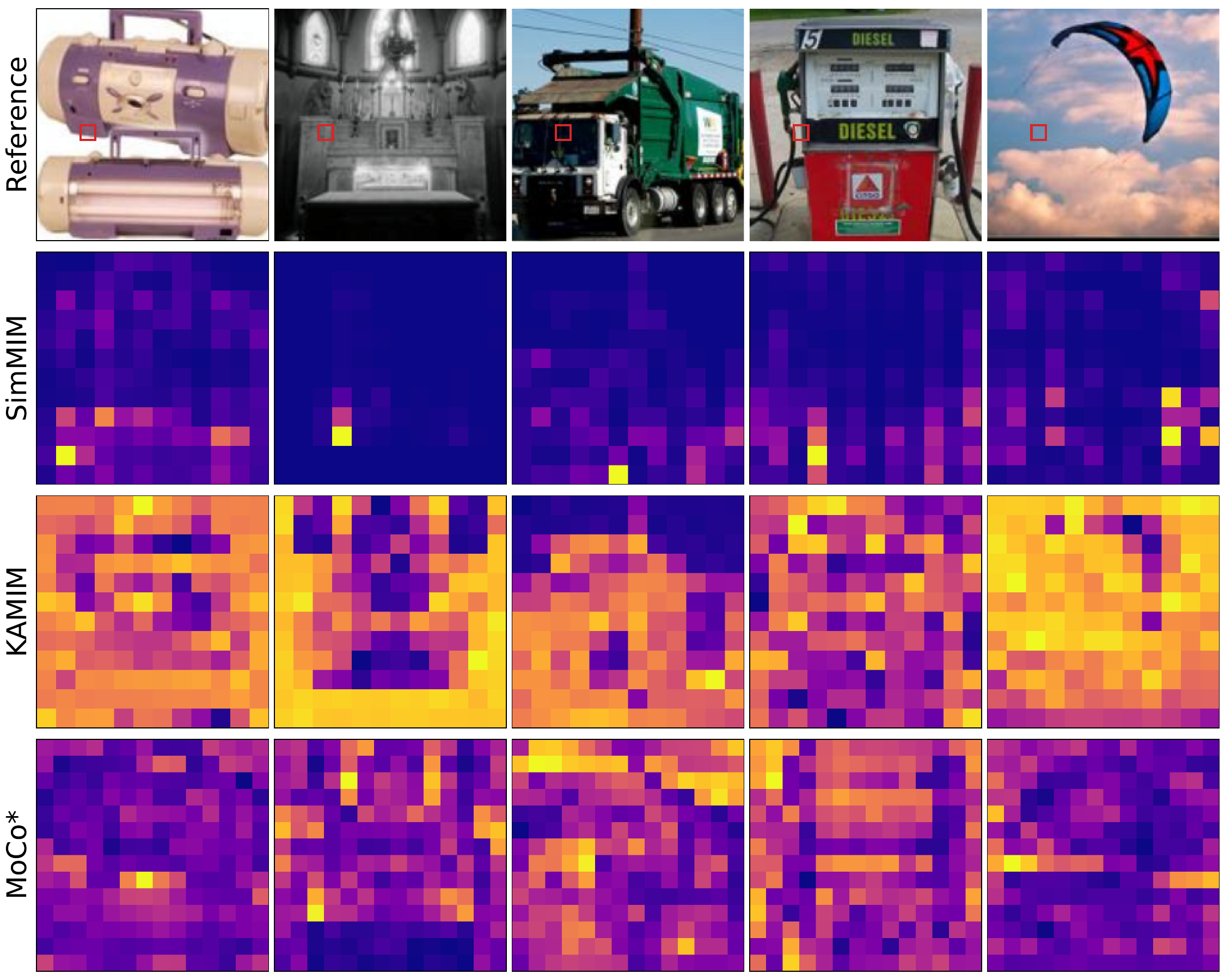}
\vspace{-0.25cm}
\caption{The attention maps for the query token (marked with a red box) for SimMIM, KAMIM, and MoCo, obtained from the last layer of a ViT-B. The upper row indicates the base image, and subsequent rows represent the attention maps for SimMIM, KAMIM, and MoCo. The attention maps for SimMIM and KAMIM have $12 \times 12$ tokens while that for MoCo has $14 \times 14$ tokens.}
\label{fig:attn_maps}
\vspace{-0.4cm}
\end{figure*}

Specifically, we observe that KAMIM behaves like contrastive learning, with higher attention distances, the ability to capture global relationships and lower normalized mean information.

\subsection{Token-level t-SNE Visualization}
We visualize the tokens of the last layer of SimMIM, KAMIM and MoCo. We see that tokens from KAMIM congregate into clusters that are grouped tighter than in SimMIM, as in \cref{fig:tsne_token_level}. While MoCo's representations remain the most separable, there is a higher degree of separability when comparing KAMIM to SimMIM, which could explain why it outperforms SimMIM in linear probing.

\subsection{Fourier Analysis of Representations}
Here, we analyze the learned representations using the magnitude of the amplitude of high-frequency signals from their Fourier-transformed feature maps as per Park and Kim \cite{park2022vision}. Higher values of $\Delta \log \text{amplitude}$ indicate a greater focus on high-frequency information.

On conducting the Fourier analysis of learned representations for each layer, as in \cref{fig:fourier_analysis}, we note that the later layers of KAMIM exploit lower-frequency signals like SimMIM and MoCo. The earlier layers of KAMIM and SimMIM exploit higher-frequency signals, while those of MoCo exploit lower-frequency signals. The plot for KAMIM also decreases monotonically, in contrast with SimMIM and MoCo's plots.

\subsection{Behaviour of Self-attention}
Here, we analyze the behavior of the self-attention mechanisms between ViT-B models trained with the three methods. In general, we observe that KAMIM has a higher attention distance, incorporates more global structures in its attention maps, and, as before, behaves more like contrastive learning than SimMIM.

\subsubsection{Attention Maps and Distances}
\begin{figure}[t!]
\centering
\includegraphics[width = 0.65\linewidth]{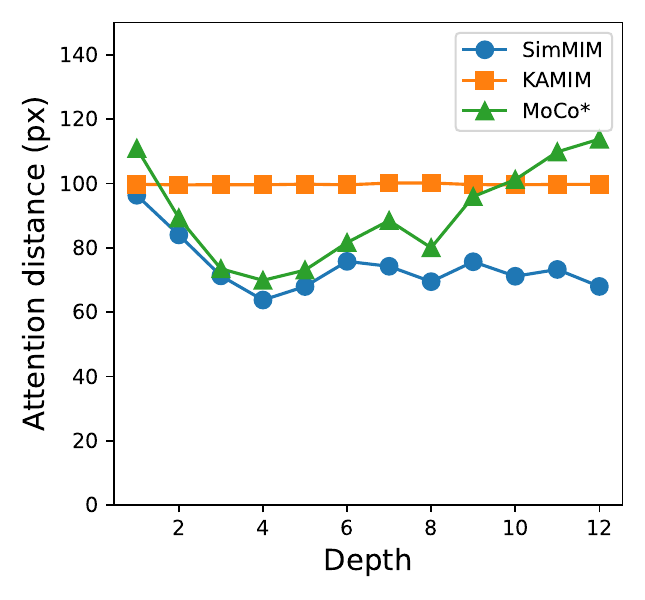}
\caption{Attention distances for SimMIM and KAMIM. The attention distance is defined as the distance between the query tokens and key tokens computed as per the weights of the self-attention. The attention distances for KAMIM ranged from 99.57 pixels to 99.97 pixels.}
\label{fig:attn_dist}
\vspace{-0.5cm}
\end{figure}

KAMIM results in more global attention maps, while SimMIM is more local. This can be seen from the attention maps in \cref{fig:attn_maps}. The attention maps are obtained from the features of the last layer of a ViT-B. KAMIM's attention maps are similar to that of MoCo, having a larger attention distance. On the other hand, SimMIM remains largely local in nature. \cref{fig:attn_dist} provides more evidence of this by showing that the attention distance for KAMIM remains largely constant at around 100 pixels, while SimMIM starts high but ends up at around 64 pixels. MoCo has a different attention distance for each layer, and it ends up being higher than KAMIM at 118 pixels.

\setcounter{figure}{7}
\begin{figure}[b]\setlength{\hfuzz}{1.1\columnwidth}
\begin{minipage}{\textwidth}
\centering
\includegraphics[width =\textwidth]{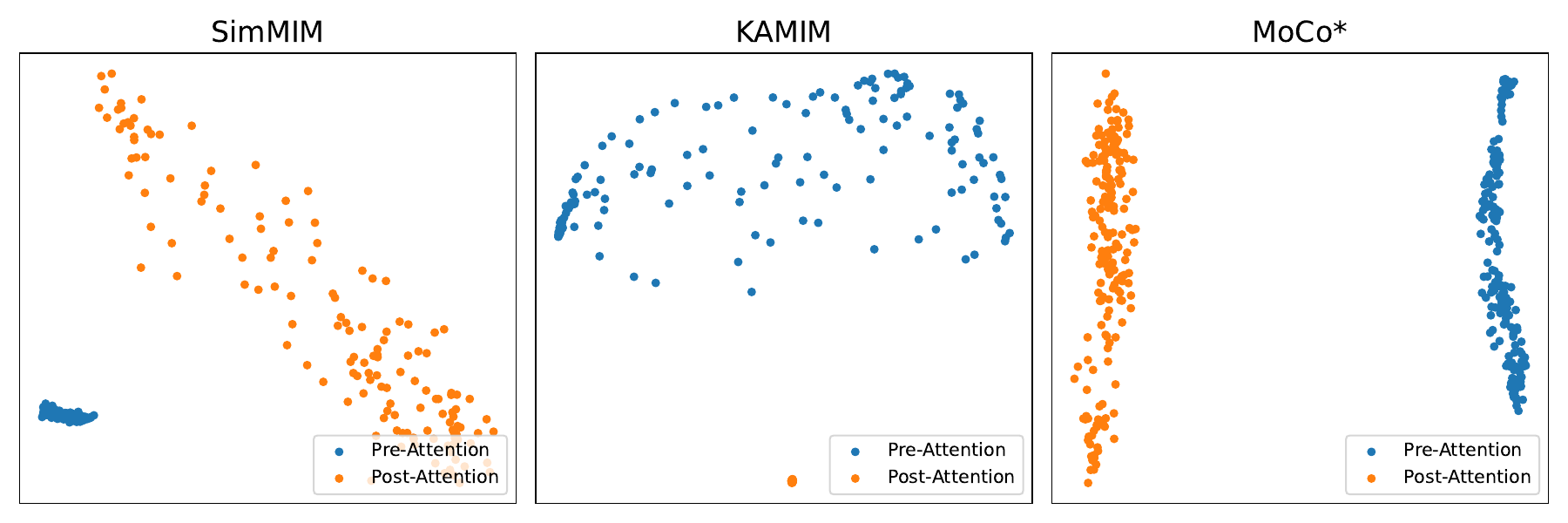}
\vspace{-0.5cm}
\caption{A plot showcasing how tokens get affected by the self-attention transformation for SimMIM and KAMIM. The blue markers are pre-attention, while the orange markers are the post-attention tokens. The embeddings were taken from the 11th layer for all plots.}
\label{fig:attn_trans_tsne}
\vspace{-0.2cm}
\end{minipage}
\end{figure}

\subsubsection{Normalized Mutual Information (NMI)}
Normalized Mutual Information (NMI) plots, shown in \cref{fig:normalized_mi}, indicate that KAMIM undergoes an attention collapse like contrastive learning at the last layer. This means that the layers collapse into homogenous distributions, and the attention maps are less dependent on the position of query tokens. While both MoCo and SimMIM show a varying NMI curve, KAMIM's plot remains largely constant and close to zero.

\subsubsection{Transformation of Tokens by Self-Attention}
We plot the embeddings before and after self-attention in \cref{fig:attn_trans_tsne}. by using PCA \cite{pearson1901liii} to visualize the tokens in 2D. We see that KAMIM closely clusters all tokens after transformation, giving more evidence for the attention-collapse phenomenon observed in \cref{fig:normalized_mi}. SimMIM and MoCo work quite differently from KAMIM. SimMIM translates each token by different amounts, while MoCo translates all tokens by roughly the same amount.

\setcounter{figure}{6}
\begin{figure}[t]
\centering
\includegraphics[width = 0.7\linewidth]{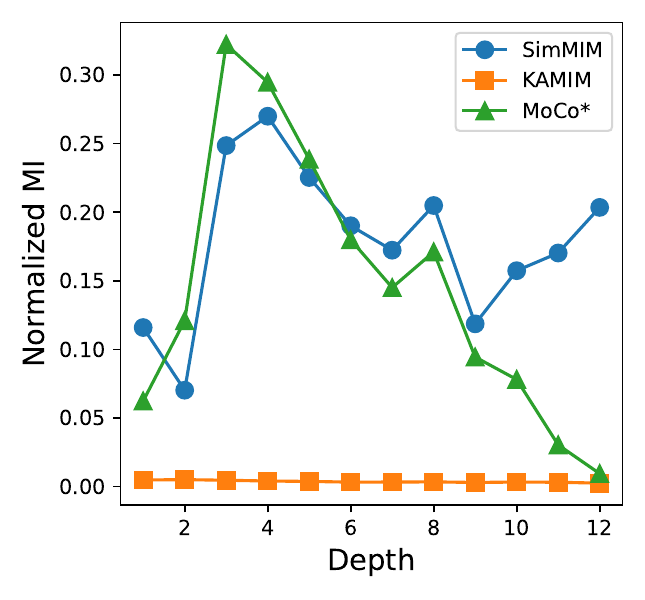}
\vspace{-0.3cm}
\caption{Plots for Normalized Mutual Information (NMI) for SimMIM and KAMIM. NMI captures the inhomogeneity of self-attentions with respect to queries. A lower NMI value shows that maps are not dependent on the location of the query, indicating a collapse of self-attention. The NMI for KAMIM ranged from 0.0024 to 0.0056.}
\label{fig:normalized_mi}
\vspace{-0.4cm}
\end{figure}

\section{Conclusion}
\label{sec:conclusion}
We propose KAMIM, a self-supervised pretraining method that builds on SimMIM by weighting the pixel-wise reconstruction by a function of the keypoint density in each patch. We observe how the use of handcrafted keypoint detectors such as FAST, ORB, and SIFT can significantly improve linear probing performance when used with SimMIM while still remaining efficient with regard to the training time. We draw parallels between KAMIM and MoCo, a method for contrastive learning through analyses of the self-attentions and representations, which helps explain KAMIM's superior linear probing performance.

\clearpage

\section*{Acknowledgement}
We acknowledge the support of the iHUB-ANUBHUTI-IIITD FOUNDATION set up under the NM-ICPS scheme of the Department of Science and Technology, India, and Google's TPU Research Cloud (TRC).

{\small
\bibliographystyle{ieee_fullname}
\bibliography{egbib}
}

\clearpage

\begin{appendices}
\section{Performance with LayerNorm}

\begin{table}[h]
\begin{tabular}{lll}
\hline
\toprule
\textbf{Model} & \textbf{SimMIM (LP)} & \textbf{KAMIM (LP)} \\ \midrule
w/o LayerNorm  & 16.12                & \textbf{33.97}               \\
w/ Layernorm   & 20.51                & \textbf{37.72}               \\ \bottomrule
\end{tabular}
\caption{Top-1 accuracies on ImageNet-1k with KAMIM and SimMIM with and without Layernorm. We evaluate linear probing performance using the embeddings of the 8th layer with ViT-B models pre-trained on ImageNet-1k.}
\label{tab:layernorm_performance}
\end{table}

We observe the impact of LayerNorm in linear probing with SimMIM and KAMIM in \Cref{tab:layernorm_performance}. We test ViT-B models pre-trained on ImageNet with an AdamW optimizer with $lr = 8e-4$, $\beta_1 = 0.9$ and $\beta_2 = 0.999$. We employ a Cosine LR scheduler for 100 epochs with 10 epochs of warmup. We then perform linear probing using the $8^{th}$ layer embeddings for the 100 epochs with 10 epochs of warmup with a Cosine LR scheduler and an AdamW optimizer that has the same values of $beta_1$ and $beta_2$ as that used for pre-training but with $lr = 5e-3$.  

We see that KAMIM leads SimMIM with and without LayerNorm. Despite this, LayerNorm increases the performance of KAMIM by $3.75\%$ and the accuracy with SimMIM by $4.39\%$.

\setcounter{figure}{8}
\begin{figure}[b]\setlength{\hfuzz}{1.1\columnwidth}
\begin{minipage}{\textwidth}
\centering
{\includegraphics[width = \textwidth]{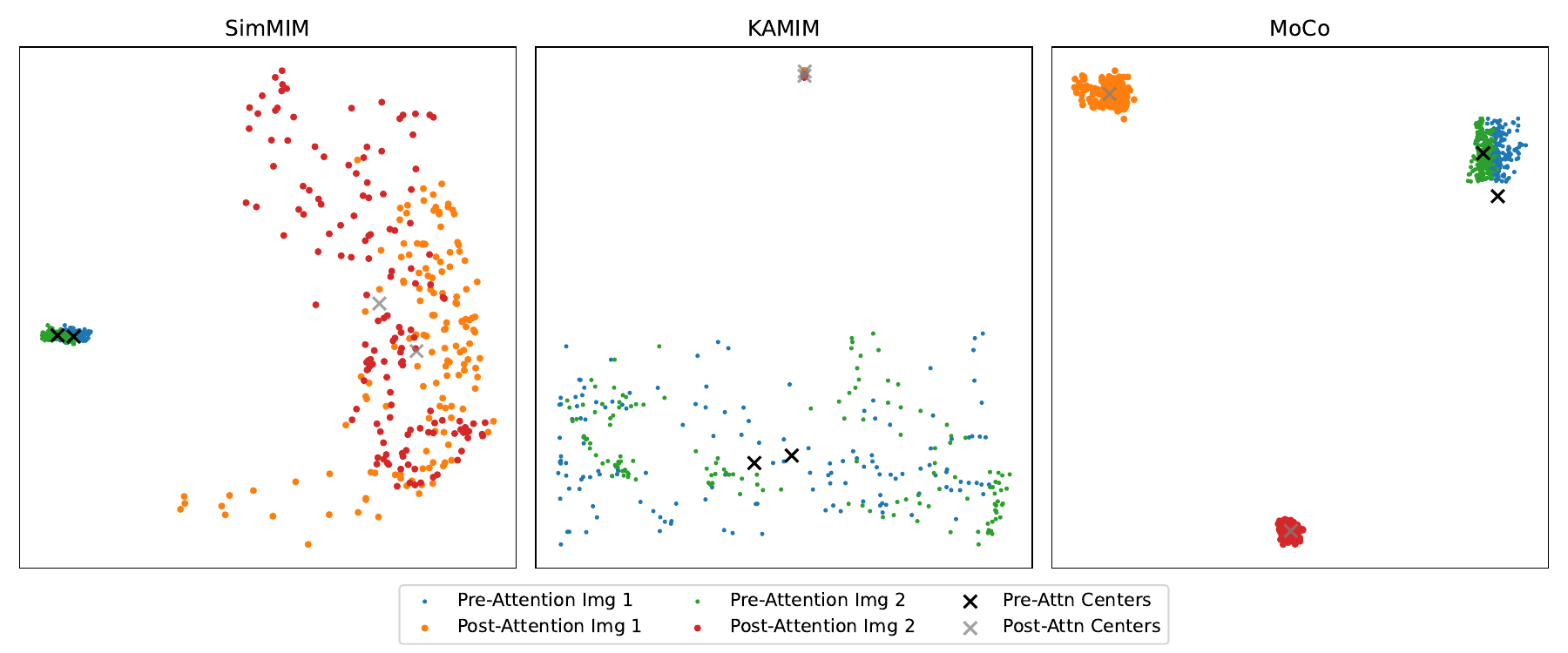}}
\caption{The effect of the self-attention on tokens of two images of distinct classes. The blue and green dots are pre-self-attention tokens of the two classes, respectively. The orange and the red dots are the post-self-attention tokens of the two respective classes. The black crosses represent the pre-self-attention centroids, and the grey crosses represent the post-self-attention centroids of tokens.}
\label{fig:pre_post_attn}
\end{minipage}
\end{figure}

\section{Transformation of Two Images After Self-Attention}

We visualize how the self-attention transformation affects tokens in SimMIM, KAMIM, and MoCo as per Park \etal in \cref{fig:pre_post_attn}. We perform Principal Component Analysis (PCA) on the tokens obtained from the last layer of ViT-B transformers to plot them on a 2D grid. We see that all three pre-training methods are drastically different in terms of their behaviors. KAMIM projects tokens to the same point, and the centroids for the two classes after the transformation lie close by. SimMIM spreads tokens from the centroids apart, but the individual tokens are intermingled. MoCo pulls apart tokens from different classes and increases the distances between them, which contributes to its excellent performance in linear probing.

\begin{figure*}[t]
\centering
{\includegraphics[width = \textwidth]{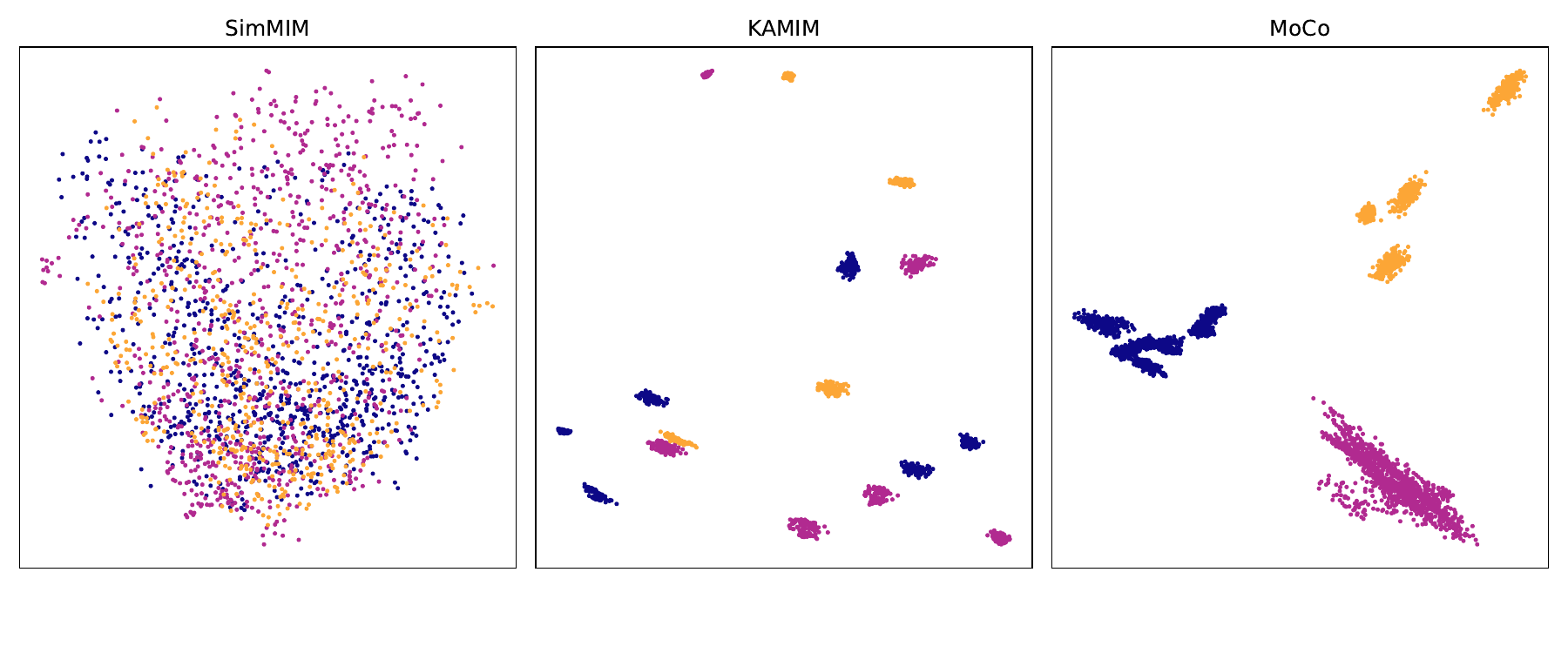}}
\caption{Plots of tokens of 16 images of 3 different classes following the self-attention transformation for SimMIM, KAMIM, and MoCo. Tokens of images from the same class have the same color. There were 6 images of class 1, 6 images of class 2, and 4 images of class 3.}
\label{fig:post_attn}
\end{figure*}

In \cref{fig:post_attn}, we see that in the case of KAMIM, the tokens after self-attention get clustered for each image, even though there is no clustering observed on a per-class level. The tokens from SimMIM get intermingled and we cannot discern clusters based on images or classes. MoCo results in separable clusters wherein images of the same class form larger clusters. This offers more separability than KAMIM, which in turn is more separable than SimMIM. This explains why KAMIM performs better than SimMIM but loses out to MoCo in linear probing.
\end{appendices}

\end{document}